\pdfoutput=1

\documentclass[11pt]{article}
\usepackage[final]{acl}

\usepackage{times}
\usepackage{latexsym}

\usepackage[T1]{fontenc}

\usepackage[utf8]{inputenc}

\usepackage{microtype}

\usepackage{inconsolata}

\usepackage{graphicx}

\usepackage{inconsolata}

\usepackage[utf8]{inputenc}
\usepackage{CJKutf8}

\usepackage{enumitem}
\usepackage{caption}
\usepackage{subcaption}
\usepackage{graphicx}
\usepackage[inkscapelatex=false]{svg}
\usepackage{booktabs}
\usepackage{multirow}
\usepackage{arydshln} 
\usepackage{tablefootnote}

\usepackage{stfloats}

\title{MMTE: Corpus and Metrics for Evaluating Machine Translation Quality of Metaphorical Language}

\author{Shun Wang\textsuperscript{1}, Ge Zhang\textsuperscript{3,5}, Han Wu\textsuperscript{4}, Tyler Loakman\textsuperscript{1}, Wenhao Huang\textsuperscript{5}, Chenghua Lin\textsuperscript{1,2}\thanks{Corresponding author}\\
\textsuperscript{1}Department of Computer Science, The University of Sheffield, UK \\
\textsuperscript{2}Department of Computer Science, The University of Manchester, UK \\
\textsuperscript{3}Department of Computer Science, University of Waterloo, Canada \\
\textsuperscript{4}School of Foreign Studies, UIBE, Beijing, China ~~~ \textsuperscript{5}01.AI, Beijing, China\\
\texttt{\{swang209, tcloakman1\}@sheffield.ac.uk},~~~
\texttt{zhangge@01.ai}\\
\texttt{wuhan@uibe.edu.cn}, ~~~ \texttt{chenghua.lin@manchester.ac.uk}
}

\begin{document}
\begin{CJK*}{UTF8}{gbsn}
{\makeatletter\acl@anonymizefalse
  \maketitle
}
\begin{abstract}
Machine Translation (MT) has developed rapidly since the release of Large Language Models and current MT evaluation is performed through comparison with reference human translations or by predicting quality scores from human-labeled data. However, these mainstream evaluation methods mainly focus on fluency and factual reliability, whilst paying little attention to figurative quality. 
In this paper, we investigate the figurative quality of MT and propose a set of human evaluation metrics focused on the translation of figurative language. We additionally present a multilingual parallel metaphor corpus generated by post-editing. 
Our evaluation protocol is designed to estimate four aspects of MT: Metaphorical Equivalence, Emotion, Authenticity, and Quality. In doing so, we observe that translations of figurative expressions display different traits from literal ones. 
\end{abstract}
\section{Introduction}
\label{sec:intro}
Metaphorical expressions are widely used in daily life for communication and vivid description, drawing attention from psycholinguistics and computational linguistics due to their key role in the cognitive and communicative functions of language \citep{wilks1978making,lakoff1980metaphors, lakoff1993contemporary}. Linguistically, a metaphor is defined as a figurative expression that uses one or more words to represent another concept within a given context, rather than taking the literal meaning of the expression \citep{fass1991met}. 
For instance, in the sentence ``\textit{The scream \underline{pierced} the night.}'', the contextual meaning of \textit{pierced} is to ``sound sharply or shrilly'', which differs from its literal meaning of ``cut or make a way through''.\textcolor{blue}{\footnote{\url{http://wordnetweb.princeton.edu/perl/webwn}}}

A significant portion (e.g. up to 20\%) of our everyday language is delivered in metaphorical terms~\cite{steen2010method}. According to \citeauthor{lakoff1980metaphors}'s study, metaphor is a type of conceptual mapping. The cognitive model, which involves reasoning about one thing in terms of another, has been shown to affect our decision-making and perception~\cite {lakoff1980metaphors,lakoff1993contemporary, boroditsky2011language}. Research also suggests that this concept-to-concept mapping is often language-agnostic, with similar mappings being feasible across different languages~ \citep{tsvetkov2014metaphor}. For instance, in the aforementioned example, the English word “\textit{pierce}” corresponds to the Chinese word “\textit{穿透}”, which literally means “pass through an object or medium” and is also used as a metaphor to indicate a sudden and sharp sound.

\begin{figure}[t]
    \centering
    \includegraphics[width=1.0\columnwidth]{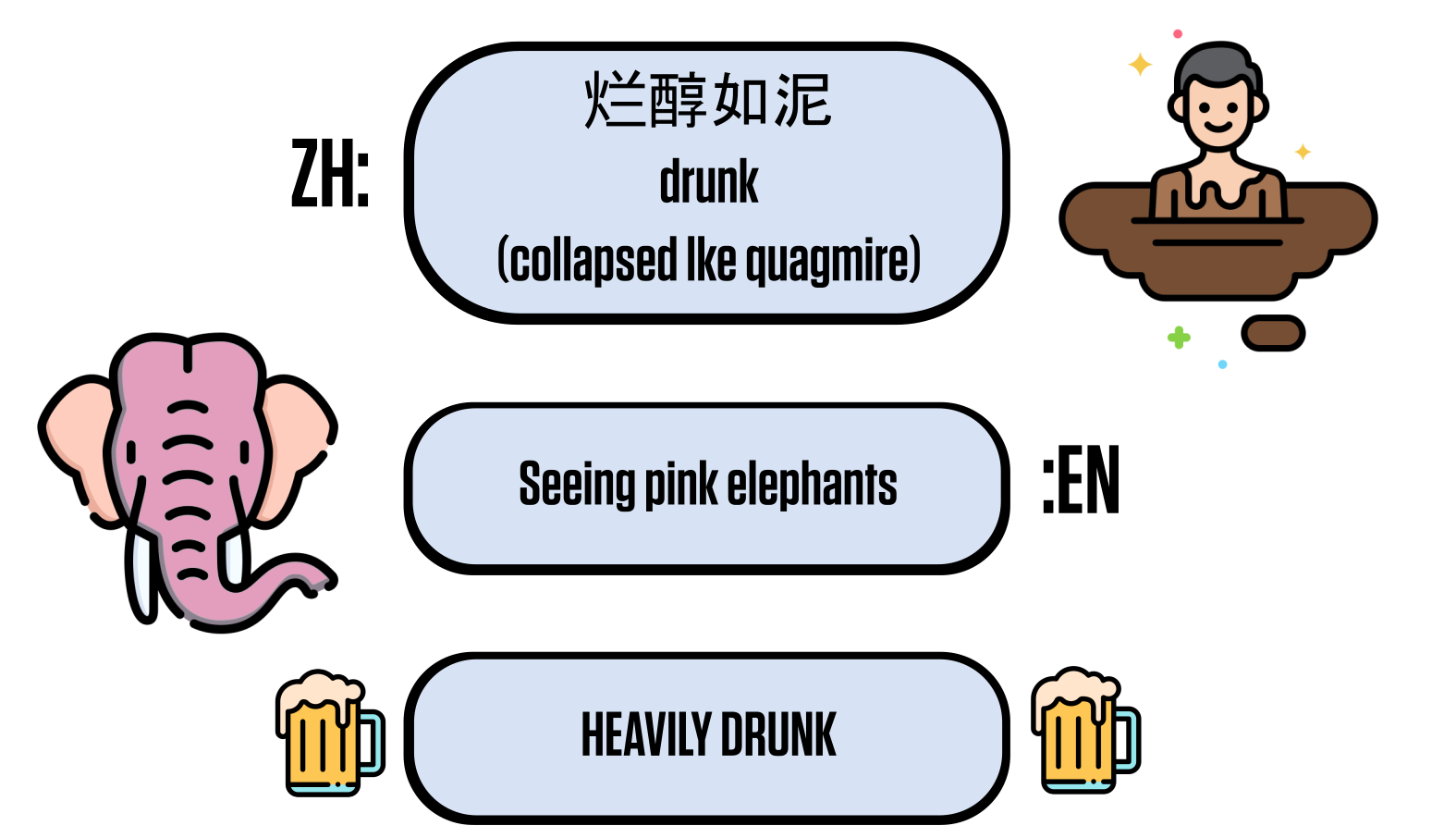}
    \caption{Chinese and English metaphorical expressions of being drunk.}
    \label{fig: drunk metaphor}
\end{figure}

However, direct translations cannot always be found in the target language due to linguistic and cultural differences. To illustrate this issue, we provide an example of metaphorical expressions of being \textit{drunk} in \autoref{fig: drunk metaphor}, where in Chinese it is common to compare drunkenness to being \textit{collapsed on the ground like quagmire}, whilst in English it is common to compare it to \textit{seeing pink elephants}. These word-sense misalignments caused by different linguistic norms are ubiquitous in practical translation applications. 

Metaphor, and especially metaphor \textit{translation}, has received increasing attention in linguistics \cite{qin2022cognitive, anvarovna2022conceptual,li-etal-2023-framebert}, and its significance has also been highlighted in various NLP tasks similar to translation, such as poem writing \cite{chakrabarty2020generating}, story generation \cite{chakrabarty2021mermaid}, and dialogue \cite{oprea2022should}.
Unfortunately, the challenge of machine translating metaphorical language remains largely unaddressed due to a scarcity of resources such as parallel data \cite{mao-etal-2018-word, gamonal-2022-descriptive}. 
To remedy this, we propose MMTE - the first systematic study of \textbf{M}etaphorical \textbf{M}achine \textbf{T}ranslation \textbf{E}valuation to explore the difficulties inherent in translating metaphorical expressions. 
Our contributions include:
\begin{itemize}[topsep=0pt,parsep=0pt]
    \setlength{\itemsep}{0pt}
    \setlength{\parsep}{0pt}
    \setlength{\parskip}{0pt}
    \item \textbf{\textit{Corpus}}: The first manually annotated multilingual metaphor translation evaluation corpus between English and Chinese/Italian.
    \item \textbf{\textit{Human Evaluation Framework}}: The first systematic human evaluation framework for metaphor translation. We also introduce rhetorical \textbf{Equivalence} for metaphorical translation Quality Estimation (QE). 
    \item \textbf{\textit{Theoretical Linguistic Foundations}}: We demonstrate the difficulties of metaphor translation from a multilingual and multi-perspective approach and provide a systematic framework for metaphor translation.
\end{itemize}

\section{Related Work} 
\textbf{Metaphor-related Tasks.} Metaphors play a crucial role in daily communication and understanding human emotion \cite{mohammad2016metaphor} and cognition \cite{tong2021recent}. 
Enhancing the quality of our understanding of metaphors has been shown to be crucial for various natural language understanding (NLU) tasks, including natural language inference (NLI) \cite{stowe2022impli}, sentiment analysis \cite{alsiyat-piao-2020-metaphorical,li-etal-2022-secret}, humour explanation \cite{mittal2022survey}, and offensive language detection \cite{tang2020categorizing}. 

Moreover, adequately conveying metaphor is a staple concern in various natural language generation (NLG) tasks, including poem writing \cite{ liu2019rhetorically, chakrabarty2020generating}, paraphrasing \cite{bizzoni-lappin-2018-predicting, stowe-etal-2021-exploring}, dialogue generation \cite{zheng2019love, oprea2022should}, and story generation \cite{chakrabarty2021mermaid}. Appropriate use of metaphor has been shown to dramatically improve user satisfaction with such systems \cite{li2022cm}, and exploring the mechanisms behind metaphor generation helps test and verify cognitive theories of how metaphors are created and used \cite{lederer-2016-finding, dankers-etal-2019-modelling}. Standalone metaphor generation has also been identified as a significant branch of creative NLG in itself \cite{stowe2021metaphor, chakrabarty2021mermaid, li2022cm,ge2023survey, shao-etal-2024-cmdag-chinese}.

Existing multilingual metaphor research \cite{mohler-etal-2014-novel, kozareva-2015-multilingual, gordon-etal-2015-high} has primarily focussed on multilingual metaphor detection and identification guided by metaphorical mappings \cite{shutova-etal-2017-multilingual}, the polarity and valence of multilingual metaphors \cite{kozareva-2015-multilingual}, and general metaphor frames \cite{gamonal2022descriptive, aghazadeh2022metaphors}.
Researchers have also explored metaphor detection and generation mechanisms in languages besides English, including Chinese \cite{chung-etal-2020-metaphoricity, li2022cm}, Malay \cite{chung2005market}, Arabic \cite{alsiyat-piao-2020-metaphorical}, and German \cite{schneider-etal-2022-metaphor}.
Metaphor translation is also a prevalent topic in linguistics \cite{pranoto2021analysis, qin2022cognitive, anvarovna2022conceptual}, whilst relatively unexplored in the machine translation domain.
Whilst some research has explored metaphor's impact on machine translation \cite{gamonal-2022-descriptive,li2024finding}, this has focussed on applying external knowledge bases to enhance cross-lingual metaphor detection. This lack of guidance from professional translators results in translations that lack linguistic nuance.

\noindent \textbf{Fine-grained Translation Quality Estimation.} Translation Quality Estimation (QE) has received increased attention, yet remains an open challenge due to resource scarcity and difficulties in handling the variation in linguistic forms and cultural norms that is inherent in metaphor \cite{vamvas-sennrich-2022-little, lu-etal-2022-learning}.
Existing translation QE work, including traditional automatic metrics~\cite{martins-etal-2017-pushing,baek-etal-2020-patquest}, encoder LM-based metrics~\cite{ranasinghe2020transquest, zheng-etal-2021-self}, and generative LLM-based metrics~\cite{kocmi2023large,lu2023error,zhao2024handcrafted}, underestimates differences caused by the cultural phenomena that underlie the use of different languages.

Although some exploratory works investigate the influence of cultural norms in the target language \cite{vela2014beyond, eo2022quak}, cross-lingual patterns \cite{zhou2020zero}, and pivot languages (i.e., an intermediary language for translation between many different languages) on machine translation QE \cite{zou2022investigating}, few works provide specific metrics to analyse how machine translation models perform on maintaining the linguistic phenomena of the target language, including culture-bound figurative description.

\noindent \textbf{Metaphor Quality Estimation.} Metaphor Quality Estimation (MQE) mainly adopts the use of human evaluation \cite{loakman-etal-2023-iron}, or directly compares the meanings of tenors (i.e., the subject of a description) and vehicles (i.e., the figurative language used to describe the tenor) \cite{li2022cm}. 
However, human evaluation of metaphor leads to difficulty in constructing fair contradistinctions between examples \cite{zayed-etal-2020-figure},
and additionally, directly comparing the meanings of tenors and vehicles ignores the reality that metaphors can be generated on the basis of various different aspects of the vehicles and tenors, which may span long stretches of text, leading to the introduction of noise from misjudgments \cite{stowe2021metaphor,wang-etal-2023-metaphor}.
\citet{miyazawa-miyao-2017-evaluation, miyazawa2019automatically} investigate manual metrics for metaphorical expression. The proposed metrics primarily focused on annotators evaluating metaphoricity by assigning one single score, therefore lacking an in-depth exploration of metaphor types and underlying principles.
\citet{distefano2024automatic} investigates adopting LLMs for metaphor scoring but only for creativity assessment.

In contrast to existing studies, MMTE proposes the first systematic human evaluation framework for performing comprehensive examinations and evaluations of metaphor translation and assessing its complexities and challenges.

\begin{figure*}[htbp]
    \centering  
    \includegraphics[width=0.9\textwidth]{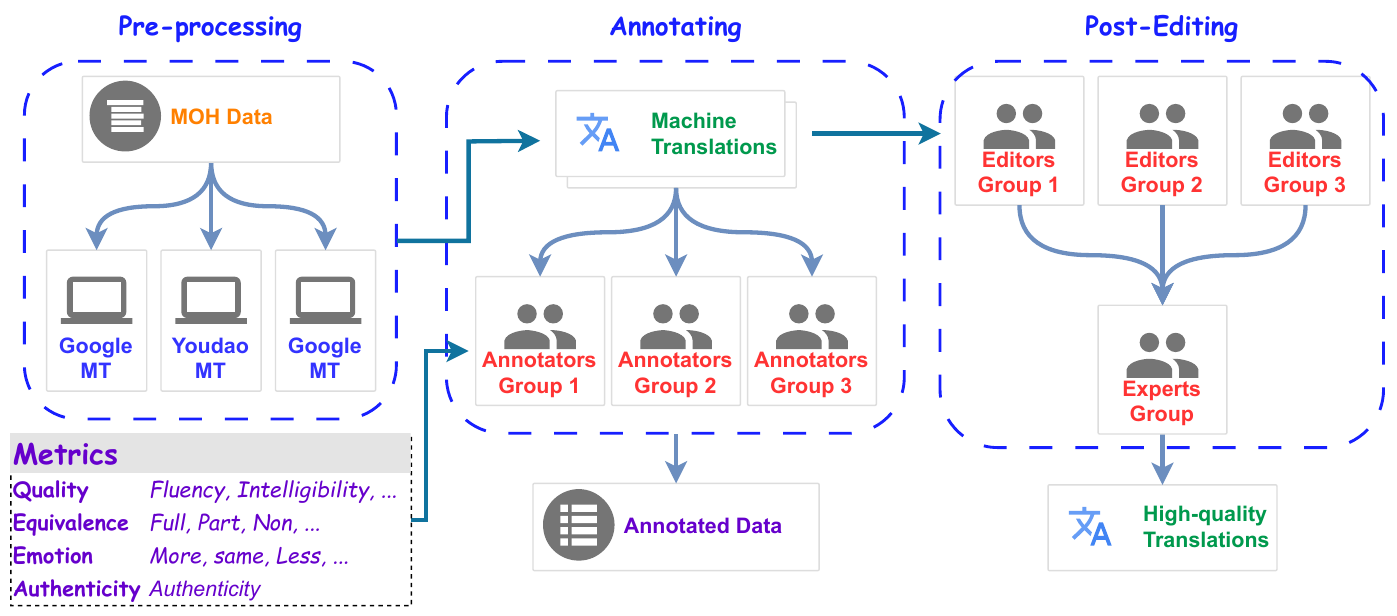}
    \caption{The dataset creation framework. By translating, annotating, and post-editing, we create a cross-lingual metaphor dataset. Specific details of these sub-steps are elaborated in Sections~\ref{sec:translating}, \ref{sec:annotating}, and \ref{sec:post-editing}, respectively.}
\label{Fig:framework}
\end{figure*}

\section{Metaphorical Translation Quality Annotation Framework}
As discussed in \S\ref{sec:intro}, metaphorical expressions are not evaluated sufficiently with current MT evaluation metrics.
To address this issue, we propose a set of novel MT evaluation metrics based on manual annotation and post-editing. 
The proposed metrics aim to provide a more accurate and insightful assessment of MT performance in handling metaphors. Our framework allows for the evaluation of MT outputs in terms of their metaphorical expressions, enabling a more comprehensive analysis of their effectiveness in capturing the nuanced meaning conveyed by such expressions, as demonstrated in \autoref{Fig:framework}.

\subsection{Initial Dataset Translation}
\label{sec:translating}
\begin{table*}[htb]
\resizebox{\textwidth}{!}{
\centering
\small
\begin{tabular}{lp{3.5cm}p{3.5cm}p{3.5cm}p{3.5cm}}
\hline
\multicolumn{1}{c}{\textbf{source instances}}            & \multicolumn{1}{c}{\textbf{google-en-zh}} & \multicolumn{1}{c}{\textbf{youdao-en-zh}} & \multicolumn{1}{c}{\textbf{opus-mt-en-zh}} & \multicolumn{1}{c}{\textbf{GPT-4o}}                      \\ \hline
The scream \underline{\textbf{pierced}} the night. & 尖叫声\underline{\textbf{划破}}黑夜。 & 尖叫声\textbf{\underline{划破}}黑夜。 & 尖叫声\underline{\textbf{刺穿}}了夜晚。& 尖叫声\underline{\textbf{刺穿}}了夜晚。 \\
The Senator \underline{\textbf{steamrollered}} the bill to defeat. & 参议员 \underline{\textbf{以 压倒性 的 方式}} 使 议案 落败 。  & 那位 参议员 \underline{\textbf{强 行}}使 该 法案 失败 。 & 参议员 把 法案 \underline{\textbf{推倒}} 了。 & 参议员 将 该法案 \textbf{\underline{压倒性 地}} 击败。              \\ \hline & \multicolumn{1}{c}{\textbf{google-en-it}}  & \multicolumn{1}{c}{\textbf{youdao-en-it}}  & \multicolumn{1}{c}{\textbf{opus-mt-en-it}}   & \multicolumn{1}{c}{\textbf{GPT-4o}}    \\ \hline
The scream \underline{\textbf{pierced}} the night. & L'urlo \underline{\textbf{squarciò}} la notte. & L’urlo \underline{\textbf{forò}} la notte. & L'urlo \underline{\textbf{ha trafitto}} la notte.  & L'urlo \underline{\textbf{ha squarciato}} la notte.\\
The Senator \underline{\textbf{steamrollered}} the bill to defeat. & Il senatore ha \underline{\textbf{schiacciato}} il disegno di legge per sconfiggerlo. & Il senatore ha \underline{\textbf{buttato via}} il disegno di legge per sconfiggerlo. & Il senatore ha \underline{\textbf{rullato}} il conto per sconfiggere. & Il senatore ha \underline{\textbf{fatto a pezzi}} il disegno di legge per sconfiggerlo. \\ \hline
\end{tabular}
}
\caption{Paired samples of source instances and their machine translations from different translation models. Target verbs are in \underline{\textbf{bold and underlined}}.}
\label{tab:parrallel_data}
\end{table*}

Due to the absence of parallel multilingual metaphor datasets, we constructed our own dataset. We employ the \textbf{MOH} dataset~\cite{mohammad2016metaphor} as our source, consisting of 315 metaphorical and 332 literal sentences sampled from WordNet \cite{miller1998wordnet}. In MMTE, \textbf{Literal} samples refer to those not containing metaphors.

We utilise four popular MT models to generate translations: the \textbf{Google Cloud Translation API}, the \textbf{Youdao Cloud Translation API}, the open-source \textbf{Helsinki-NLP/opus-mt} model from Hugging Face, and \textbf{GPT-4o} to translate English source data into \textbf{Chinese} and \textbf{Italian}, enabling us to explore and compare the treatment of metaphors in two languages with distinct characteristics. \autoref{tab:parrallel_data} presents example metaphors paired with their translations in the two target languages. Additional information regarding preprocessing is presented in Appendix~\ref{Sec:preprocess}.

\subsection{Metaphor Annotation Criteria}
\label{sec:annotating}
Our annotation protocol involves comparing translations with their source sentences. 
We hire 18 linguistics majors who are native speakers of the target languages to annotate and post-edit 647 English-Chinese (EN-ZH) and English-Italian (EN-IT) translations, with each sample being annotated by 3 individuals. Professional translators cross-checked the results, resolved disagreements in meetings, and recorded final decisions. Additional details are in Appendix ~\ref{Sec:annotation_setup}. The source instances and their corresponding translations are systematically annotated based on four criteria to evaluate translation quality: Quality, Metaphorical Equivalence, Emotion, and Authenticity. These criteria are outlined and further broken down as follows. 

\noindent\textbf{Quality.} To estimate the quality of the translation, we adopt criteria inspired by several existing human assessment methods for MT \cite{ carroll1966experiment,church1993good, white1994arpa} and consider three primary aspects of quality, including Fluency, Intelligibility, and Fidelity. Detailed definitions are presented in Appendix~\ref{Sec:guideline}.

\begin{table*}[htb]
\resizebox{\textwidth}{!}{
\centering
\small
\begin{tabular}{lp{7cm}p{7cm}}
\toprule
\multicolumn{1}{c}{\textbf{Equivalence}}            & \multicolumn{1}{c}{\textbf{Source}}                                    & \multicolumn{1}{c}{\textbf{Target}}                       \\ \hline
\multirow{2}*{\textbf{Full}} & The White House \underline{\textbf{sits}} on Pennsylvania Avenue. &   白宫\underline{\textbf{坐落}}在宾夕法尼亚大道上。 \\
& The ex-slave \underline{\textbf{tasted}} freedom shortly before she died. &  l'ex schiava ha \underline{\textbf{assaporato}} la libertà poco prima di morire.\\ \hline
\multirow{2}*{\textbf{Part}} & \underline{\textbf{Wallow}} in your success! &  \underline{\textbf{沉浸}}在你的成功中吧！\\
& My personal feelings \underline{\textbf{color}} my judement in this case. & i miei sentimenti personali \underline{\textbf{offuscano}} il mio giudizio in questo caso.
\\ \hline
\multirow{2}*{\textbf{Non}} & This drug will \underline{\textbf{sharpen}} your vision. & 这药能\underline{\textbf{改善}}你的视力。\\
 & Fire had \underline{\textbf{devoured}} our home. & l'incendio \underline{\textbf{distrusse}} la casa. \\
\bottomrule
\end{tabular}
}
\caption{Instances of various Equivalence types in metaphor translation. \textit{Full} refers to the same literal and contextual meanings; \textit{Part} means similar contextual meanings and different literal meanings while both being metaphorical; and \textit{Non} means similar contextual meanings and different literal meanings with the translation being non-metaphorical.}
\label{tab:equivalence}
\end{table*}

\noindent\textbf{Equivalence.} To ascertain how metaphors impact MT, we propose Equivalence to describe how figurative expressions are translated into another language based on two features: 1) \textit{How the meanings of the source and target are conveyed} 2) \textit{Whether or not the translation is still figurative}. By comparing source texts and translations, annotators are asked to determine to what extent the target word is Equivalent in figuration. The annotators label the translation using a set of five distinct tags, encompassing three types of Equivalence and two types of Mistranslation. We elucidate the types of Equivalence in \autoref{tab:equivalence}, based on the following definitions: 
\begin{itemize}[topsep=0pt,parsep=0pt]
    \setlength{\itemsep}{0pt}
    \setlength{\parsep}{0pt}
    \setlength{\parskip}{0pt}
    \item \textbf{\textit{Full-Equivalence}}: When comparing the source and translation, both the literal meanings and the contextual meanings of the target word are the same.
    \item \textbf{\textit{Part-Equivalence}}: When comparing the source and translation, only the contextual meanings of the target word are similar. The literal meaning of the target word between the source and translation is different, but they are both metaphorical.
    \item \textbf{\textit{Non-Equivalence}}: When comparing the source and translation, only the contextual meanings of the target word are similar. However, the translation is a non-metaphorical expression, making the literal meaning of the target words between the source and translation different.
\end{itemize}
We also identify two types of mistranslation:
\begin{itemize}[topsep=0pt,parsep=0pt]
    \setlength{\itemsep}{0pt}
    \setlength{\parsep}{0pt}
    \setlength{\parskip}{0pt}
    \item \textbf{\textit{Misunderstanding}}: When the literal meaning of the target word in the source text and translation are similar, but the translation fails to convey the contextual meaning of the target word in the source language.
    \item \textbf{\textit{Error}}: When the target word is mistranslated, meaning that not only the contextual meanings are different between the source and translation, but their literal meanings also differ.
\end{itemize}
If the source instances are non-metaphorical expressions, annotators are instructed to only classify the translations into three categories: \textbf{\textit{Literal}}, \textbf{\textit{Metaphorical}}, and \textbf{\textit{Error}}.
The non-metaphorical portion of the data is used for subsequent comparisons with the metaphorical instances.

\noindent\textbf{Emotion.} Inspired by \citet{mohammad2016metaphor}, we incorporate an analysis of emotion to investigate whether metaphorical expressions in translations convey additional emotional information compared to non-metaphorical expressions.
By comparing a source sentence and its translation, the annotators determine to what extent the target word and its translation convey different amounts of emotion. 
There are four labels to judge emotion: \textbf{\textit{Zero}}, \textbf{\textit{Less}}, \textbf{\textit{Same}}, and \textbf{\textit{More}}, separately representing that the target word in the source context conveys no emotion, or that the target word in the translation conveys less, the same, or more emotion than the target word in the source sentence.

\noindent\textbf{Authenticity.} Authenticity is an extension of existing criteria \cite{doyon1999task}, evaluating: \textit{To what extent the translated metaphor reads like standard, well-edited language, such that the metaphor would be understood by a native speaker of the target language.}
The annotators are asked to judge all aforementioned criteria on a 5-point Likert scale~\cite{likert1932technique}. 

\subsection{Post-Editing}
\label{sec:post-editing}
Due to the requirement for gold references by automatic evaluation algorithms like BLEU \cite{papineni2002bleu} and ROUGE \cite{lin2004rouge}, we introduce a post-editing method to modify the translation results of four MT models to generate a gold standard translation reference, as is common practice
\cite{senez1998post,allen2003post,somers2003computers}. 
Three groups of annotators, who are native speakers of each target language, are asked to post-edit the translations, resulting in three groups of human-edited translations for both Chinese and Italian. 
Finally, a panel of expert translators perform final filtering to select the best quality edited translation as the gold reference. Additional details regarding the human annotation process are presented in Appendix~\ref{Sec:guideline}.


We employ several quality control methods to ensure the quality of the dataset obtained through post-editing the machine translations. Annotators compare four different translations, selecting high-quality ones or modifying low-quality ones to provide a reference translation, including translations of both metaphorical and non-metaphorical language. Three separate annotator groups work on each sample. An expert panel of translators then reviewes and refines the selections. Annotators also mark the positions of target words during alignment to avoid issues in word-level processing. The final dataset includes aligned English, Chinese, and Italian translations, with 315 metaphorical and 332 literal instances per language, totalling over 1900 instances.

\subsection{Automatic Metrics for Translation Quality}
We introduce several automatic metrics to evaluate the quality of translations, which are described below.

\noindent\textbf{BLEU/ROUGE.} We use BLEU \cite{papineni2002bleu} and ROUGE \cite{lin2004rouge} as part of the automatic evaluation metrics, using the selected human-edited translations as our gold standard references. 

\noindent\textbf{BERTScore.} We use BERTScore \cite{zhang2019bertscore} as a cross-lingual translation evaluation metric to automatically evaluate translations without a target-language reference, due to BERTScore being shown to be effective in cross-lingual settings \cite{song2021sentsim}.

\noindent\textbf{GPT score.} We also employ GPT-4o\footnote{gpt-4o-2024-05-13 \url{https://platform.openai.com/}} as an annotator to score the translation results using the same scoring criteria as human annotators.

\section{Results}
\label{sec:discussion}

\begin{table*}[htpb]
  \centering
   \small
   \resizebox{\linewidth}{!}{
    \begin{tabular}{cccccccccccc}
    \toprule
     \multicolumn{2}{c}{\multirow{2}{*}{\textbf{EN-ZH}}} & \multicolumn{5}{c}{\textbf{Manual Evaluation Metrics}} & \multicolumn{5}{c}{\textbf{Automatic Evaluation Metrics}} \\
     \cmidrule(lr){3-7} \cmidrule(lr){8-12} 
     & & \multicolumn{1}{c}{Fluency} & \multicolumn{1}{c}{Intelligibility} & \multicolumn{1}{c}{Fidelity} & \multicolumn{1}{c}{Authenticity} & \multicolumn{1}{c}{Overall} & \multicolumn{1}{c}{BLEU1} & \multicolumn{1}{c}{BLEU4} & \multicolumn{1}{c}{Rouge-L} & \multicolumn{1}{c}{BERTScore} & \multicolumn{1}{c}{GPT-4o}\\
    \midrule
    Google & Metaphorical   &4.47  &4.31  &4.25  &4.12 &4.34  &0.58  &0.20  &0.62  &0.765 &4.44\\
    Google & Metaphorical (full)   & 4.75 & 4.73 & 4.72 & 4.64 & 4.71 & 0.52 & 0.20 & 0.78 &0.766 &4.75 \\
    Google & Literal   &4.53  &4.55  &4.53  &4.49 &4.54 &0.73  &0.38  &0.76  &0.768 &4.67 \\ \hdashline[2pt/5pt]
    Opus & Metaphorical   &3.87  &3.52  &3.39  &3.22 &3.59  &0.49  &0.10  &0.53  &0.737 &3.56  \\
    Opus & Metaphorical (full)   &4.40  &4.32  &4.32  &4.25 &4.32  &0.44  &0.13  &0.65  &0.735 &4.14 \\
    Opus & Literal   &3.93 &3.80  &3.74  &3.75 &3.82 &0.49  &0.14  &0.54  &0.732 &3.77   \\ \hdashline[2pt/5pt]
    Youdao & Metaphorical   &4.67  &4.59  &4.53  &4.53 &4.60  &0.64  &0.26  &0.67  &0.759 &4.64  \\
    Youdao & Metaphorical (full)   &4.82  &4.81  &4.80  &4.85 &4.82  &0.53  &0.23  &0.82  &0.764 &4.74  \\
    Youdao & Literal   &4.66   &4.67  &4.65  &4.62 &4.66  &0.80  &0.57  &0.83  &0.766 &4.74 \\
    \hdashline[2pt/5pt]
    GPT-4o & Metaphorical  & 4.05 & 4.25 & 4.35 & 4.05 & 4.17 & 0.58 & 0.26 & 0.60 & 0.764 & 4.69 \\
    GPT-4o & Metaphorical (full)   & 4.59 & 4.32 & 4.62 & 4.59 & 4.53 & 0.64 & 0.30 & 0.68 & 0.765 & 4.87  \\
    GPT-4o & Literal   & 4.54 & 4.54 & 4.17 & 4.22 & 4.37 & 0.59 & 0.29 & 0.64 & 0.761 & 4.90 \\
    \midrule \midrule
    \multicolumn{2}{c}{\textbf{EN-IT}} & \multicolumn{9}{c}{ } \\
    \midrule \midrule
    Google & Metaphorical   &4.57  &4.46  &4.30  &4.32 &4.44  &0.50  &0.22  &0.60  &0.811 &4.51\\
    Google & Metaphorical (full)   &4.78  &4.77  &4.72  &4.63 &4.73  &0.65  &0.42  &0.74  &0.811 &4.55\\
    Google & Literal   &4.77  &4.68  &4.58  &4.67  &4.68 &0.68  &0.47  &0.74  &0.807  &4.68   \\ \hdashline[2pt/5pt]
    Opus & Metaphorical   &4.45  &4.29  &4.14  &4.16  &4.29  &0.48  &0.19  &0.58  &0.808 &4.06 \\
    Opus & Metaphorical (full)   &4.78  &4.77  &4.74  &4.63  &4.73  &0.64  &0.42  &0.73  &0.809 &4.45\\
    Opus & Literal   &4.65  &4.53  &4.45  &4.52 &4.54  &0.65  &0.43  &0.71  &0.803 &4.29    \\ \hdashline[2pt/5pt]
    Youdao & Metaphorical   &4.36  &4.16  &3.96  &4.04 &4.16 &0.45  &0.17  &0.54  &0.805 &3.95 \\
    Youdao & Metaphorical (full)   &4.73  &4.73  &4.67  &4.56 &4.67 &0.61  &0.38  &0.69  & 0.801 &4.34 \\
    Youdao & Literal   &4.53  &4.42 &4.29  &4.38 &4.41  &0.58  &0.30  &0.64  &0.799 &4.13   \\
    \hdashline[2pt/5pt]
    GPT-4o & Metaphorical   & 4.41 & 4.34 & 4.14 & 4.25 & 4.28 & 0.52 & 0.24 & 0.60 & 0.812 & 4.53 \\
    GPT-4o & Metaphorical (full)   & 4.50 & 4.60 & 4.64 & 4.55 & 4.57 & 0.59 & 0.27 & 0.67 & 0.810 & 4.85  \\
    GPT-4o & Literal   & 4.59 & 4.55 & 4.50 & 4.55 & 4.55 & 0.55 & 0.26 & 0.65 & 0.811 & 4.81 \\
    \bottomrule
    \end{tabular}
    \vspace{-4mm}
    }
  \caption{Metaphorical and literal expression evaluation averages. \textbf{Manual Evaluation Metrics} and \textbf{GPT} employ a 5-point scale to assess the quality and characteristics of expressions, whilst \textbf{Automatic Evaluation Metrics} provide scores ranging 0-1. \textit{Metaphorical (full)} refers to translations annotated as having full-equivalence.}
  \label{tab: comprehensive_evaluation}
\end{table*}


We first conduct a comprehensive comparative analysis of the performance of MT on metaphorical and literal expressions based on both manual and automatic evaluation scores in \S\ref{sec:discussion}.
Specifically, by analysing the distribution of labels from the fine-grained human evaluation protocol, we verify that metaphor translation is more challenging than literal translation in \S~\ref{sec:overall}. 
Moreover, we examine the correlations between the suggested fine-grained human evaluation protocol in \S\ref{sec:llm-assess}, the correlations between Emotional and Metaphorical Expressions in \S\ref{Sec: Correlation_Emotion_Metaphor}, and the crucial role of Metaphor Equivalence in metaphor translation Quality Estimation (QE) in \S\ref{Sec: Equality Impact}. Additionally, we analyse the translation quality between typologically different languages in \S\ref{sec:language_distance}.
We also provide a case study indicating that translating between more typologically distant language pairs is harder, by comparing EN-ZH and EN-IT pairs in Appendix \ref{Sec: Language Genetic}.

\begin{figure}[tb]
    \centering
    \includegraphics[width=\columnwidth]{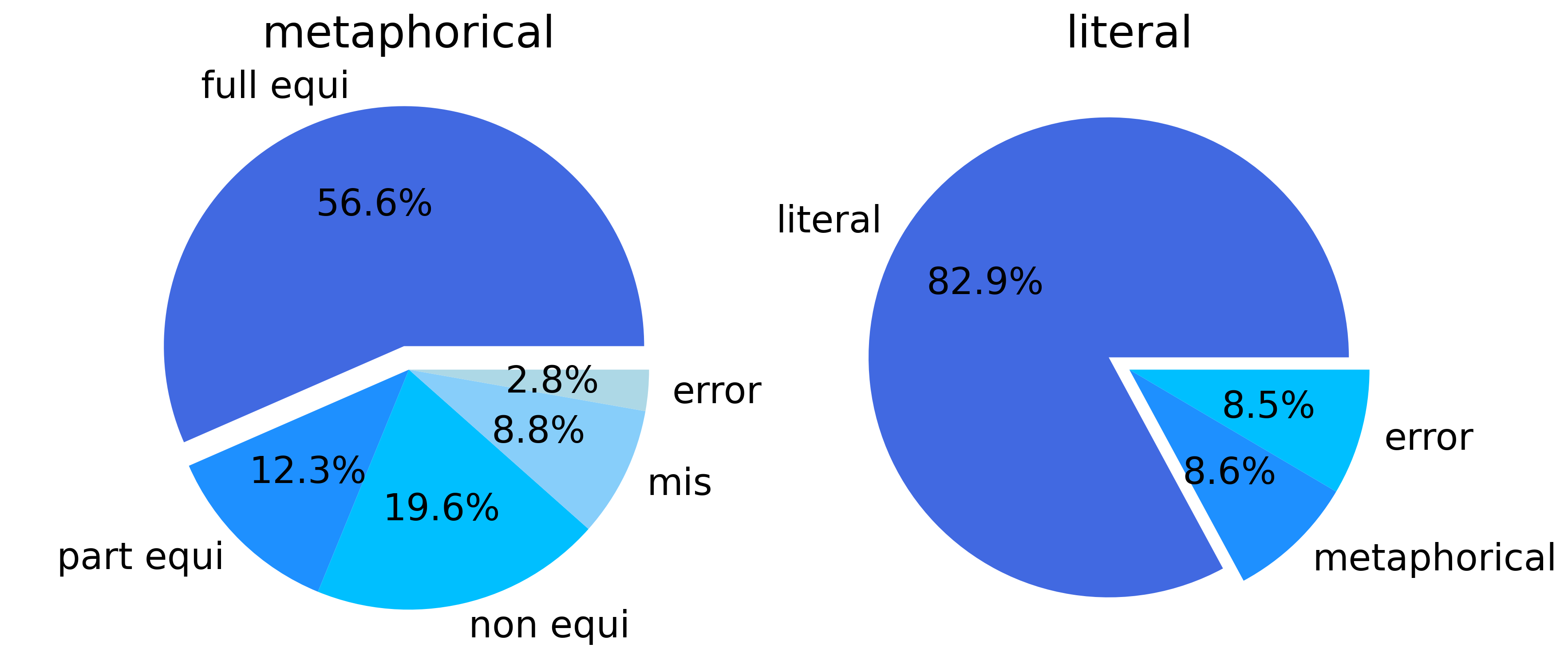}
    \caption{Equivalence distributions of metaphorical and literal expression translations from annotators. \textit{non equi}, \textit{part equi}, and \textit{full equi} refer to non-, part-, and full- equivalence, respectively. \textit{mis} denotes mistranslation.}
    \label{fig:equivalence_num}
\end{figure}

 \subsection{Metaphorical vs. Literal Translation}
\label{sec:overall}
As shown in \autoref{fig:equivalence_num}, the analysis of equivalence labels in metaphorical and literal expression translations highlights the varying degrees of equivalence and accuracy in translating metaphorical and literal expressions. 
Approximately 20\% of metaphorical expressions are found to be translated without proper correspondence to the intended metaphorical meaning \textit{(non-equi}). 
Furthermore, more than 10\% of metaphorical translations exhibit a failure to comprehend the intended metaphor or contain mistakes or inaccuracies (\textit{mis} and \textit{error}).
These results emphasise the challenges associated with translating metaphorical expressions.

\autoref{tab: comprehensive_evaluation} presents scores from manual and automatic evaluations to compare the translation of metaphorical and literal expressions.
It can be seen that translating metaphorical expressions poses greater difficulty compared to translating literal expressions.
In both EN-ZH and EN-IT translation, the metaphorical expression translations generally obtained lower scores in all Manual Evaluation Metrics and GPT-4o compared to translations of literal expressions from the same MT system. Our automatic evaluation metrics also support this observation, with lower scores from BLEU1, BLEU4, and Rouge-L for metaphorical expression translations, suggesting a reduced level of similarity and alignment with reference translations.

Most importantly, we separately calculate the evaluation scores for full-equivalence translations. The results show that when metaphors are translated faithfully, their scores are significantly higher. This demonstrates that although translating metaphors is a challenging task, achieving the correct form of translation often results in more satisfactory outcomes, therefore highlighting the importance of having comprehensive translation evaluation metrics.

BERTScore struggles to distinguish the performance between metaphorical and literal translations. 
This limitation may be due to the methods relying on contextual embeddings and cosine similarity struggles to capture the subtle semantic differences inherent in metaphorical language. 
This highlights the need for specialised evaluation tailored to the complexities of metaphor.


\subsection{LLMs Equivalence Assessment}
\label{sec:llm-assess}
\begin{table}[]
\centering
\small
{
\begin{tabular}{lccc}
\toprule
  &\textbf{GPT-3.5\tablefootnote{gpt-3.5-turbo-0125 \url{https://platform.openai.com/}}} &\textbf{GPT-4o\tablefootnote{gpt-4o-2024-05-13 \url{https://platform.openai.com/}}} & \textbf{Gemini Pro\tablefootnote{gemini-1.0-pro-001 \url{https://cloud.google.com/}}} \\ \hline
\textbf{EN-IT full} & 86.0   & 86.7 & 85.7              \\
\textbf{EN-IT others} &  92.5  & 94.0  & 91.7              \\
\textbf{EN-ZH full} & 76.2   & 76.5 & 74.4         \\ 
\textbf{EN-ZH others} & 84.1   & 86.3 & 84.7         \\ \bottomrule     
\end{tabular}
}
\caption{Accuracy of LLMs in classifying metaphor equivalence when compared to human annotations. \textit{full} refers to translations annotated as having full equivalence, whilst \textit{others} refers to translations as having non- or part- equivalence.}
\label{tab:acc_llms}
\end{table}

As shown in \autoref{tab:acc_llms}, we employ LLMs to annotate the equivalence of metaphor translations and compare the results with the human-annotated reference data. 
The LLM-based evaluation results demonstrate a high level of consistency with human annotators.
Moreover, we task LLMs with providing explanations for their annotations, offering insights into their interpretation of metaphorical content across different languages.
For instance, consider the sentence pair: EN: "She \underline{swallowed} the last words of her speech" and ZH: "她\underline{咽下}了最后几句话." Here, "咽下" is a translation with full-equivalence. The explanation from GPT-4 is as follows:
"\textit{Both in the source sentence and the translation, 'swallowed' and '咽下' are used metaphorically to mean that she did not say the last words of her speech. The literal meanings of 'swallow' and '咽下' are also the same, referring to the action of making food or drink go from your mouth down through your throat and into your stomach.}"
Detailed examples of these explanations can be found in Appendix~\ref{Sec:A_Explanation}. 
This comparison reveals that LLMs can effectively complement human efforts, providing reliable and insightful evaluations that are crucial for high-quality translation assessments at scale.
\begin{figure}[h]
    \centering
    \includegraphics[width=\columnwidth]{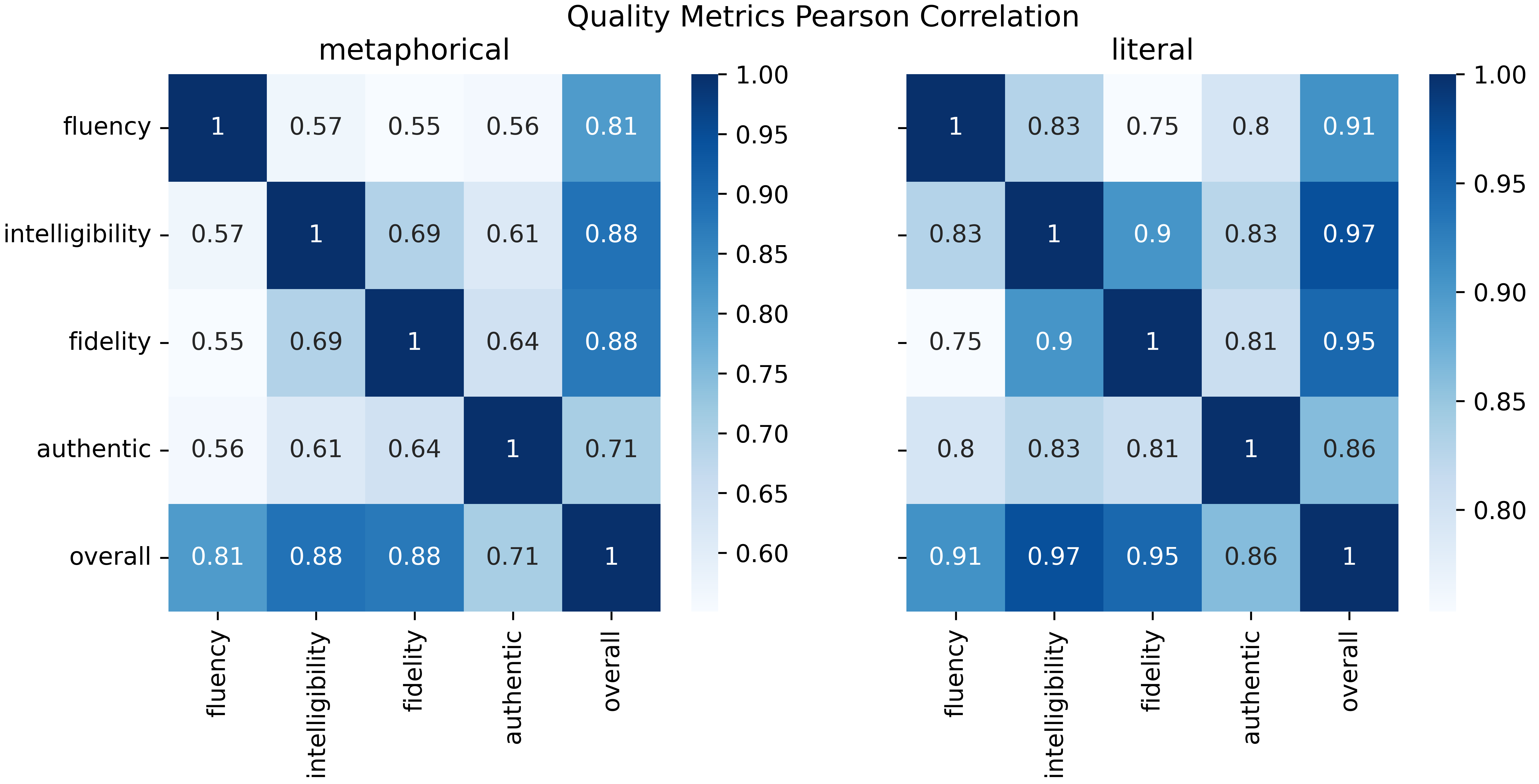}
    \caption{Pearson correlation heatmap of manual evaluation quality.}
    \label{fig:metrics_correlation}
\end{figure}

\subsection{Correlation Analysis of Fine-grained Human Evaluation Metrics}

\autoref{fig:metrics_correlation} shows the fine-grained correlations between human evaluation metrics by calculating pairwise Pearson correlations between the criteria of Fluency, Intelligibility, Fidelity, Authenticity, and Overall Score.
Firstly, we observe that the Pearson correlations between each pair are all in the interval between 0.55 and 0.65, indicating a relatively low but positive correlation among them. 
This observation verifies that Fluency, Intelligibility, Fidelity, and Authenticity represent independent aspects of metaphor translation quality estimation. 
Secondly, we observe that Fluency, Intelligibility, and Fidelity are all highly positively correlated to the Overall Score, indicating that all three elements of metaphor quality evaluation are paramount in the estimation of overall quality.

\subsection{Correlation Analysis of Emotion and Equivalence Metaphor}
\begin{figure}[!h]
    \centering
    \includegraphics[width=\columnwidth]{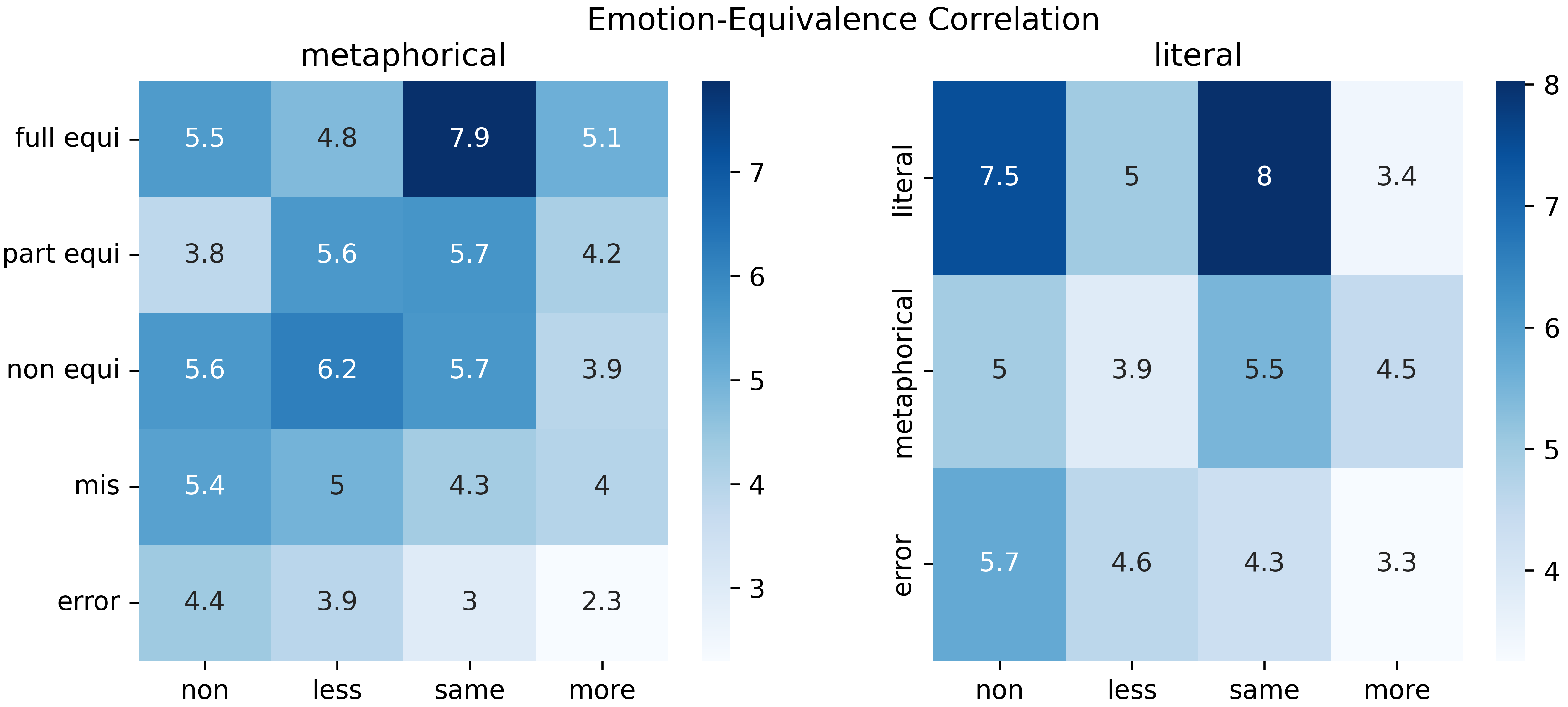}
    \caption{Emotion-Equivalence correlation heatmap based on co-occurrences.}
    \label{fig:emotion_correlation}
\end{figure}
\label{Sec: Correlation_Emotion_Metaphor}
We investigate the correlation between how much emotion the translated version retains and how figurative expressions are translated in \autoref{fig:emotion_correlation}, which presents a correlation heatmap based on a logarithmic function of the number of co-occurrences of Emotion and Equivalence defined in \S\ref{sec:annotating}. 
It is noticeable that emotion levels perceived by annotators tend to remain constant if the original metaphorical expression is translated to a fully equivalent version, and the original literal expression is translated to a literal version.  
This observation indicates that maintaining the figurative status translations is a reasonable strategy for keeping the emotional expression authentic. For example, the metaphorical expressions "\textit{swallow the sentence}" and "\textit{咽下这句话}" both convey reluctance, whilst the Chinese literal translation "\textit{没说这句话}" does not.
We also observe that non-equivalent translations tend to keep little of the emotion contained in the original metaphorical expressions. In contrast, fully equivalent and part-equivalent translations show weaker, yet similar trends.
This finding reveals the difficulty in maintaining emotion through the translation procedure and demonstrates that equivalence and whether the translation is figurative are essential for maintaining levels of emotion.

\subsection{Impact of Metaphor Equivalence}
\label{Sec: Equality Impact}

\begin{figure}[!h]
    \centering
    \includegraphics[width=\columnwidth]{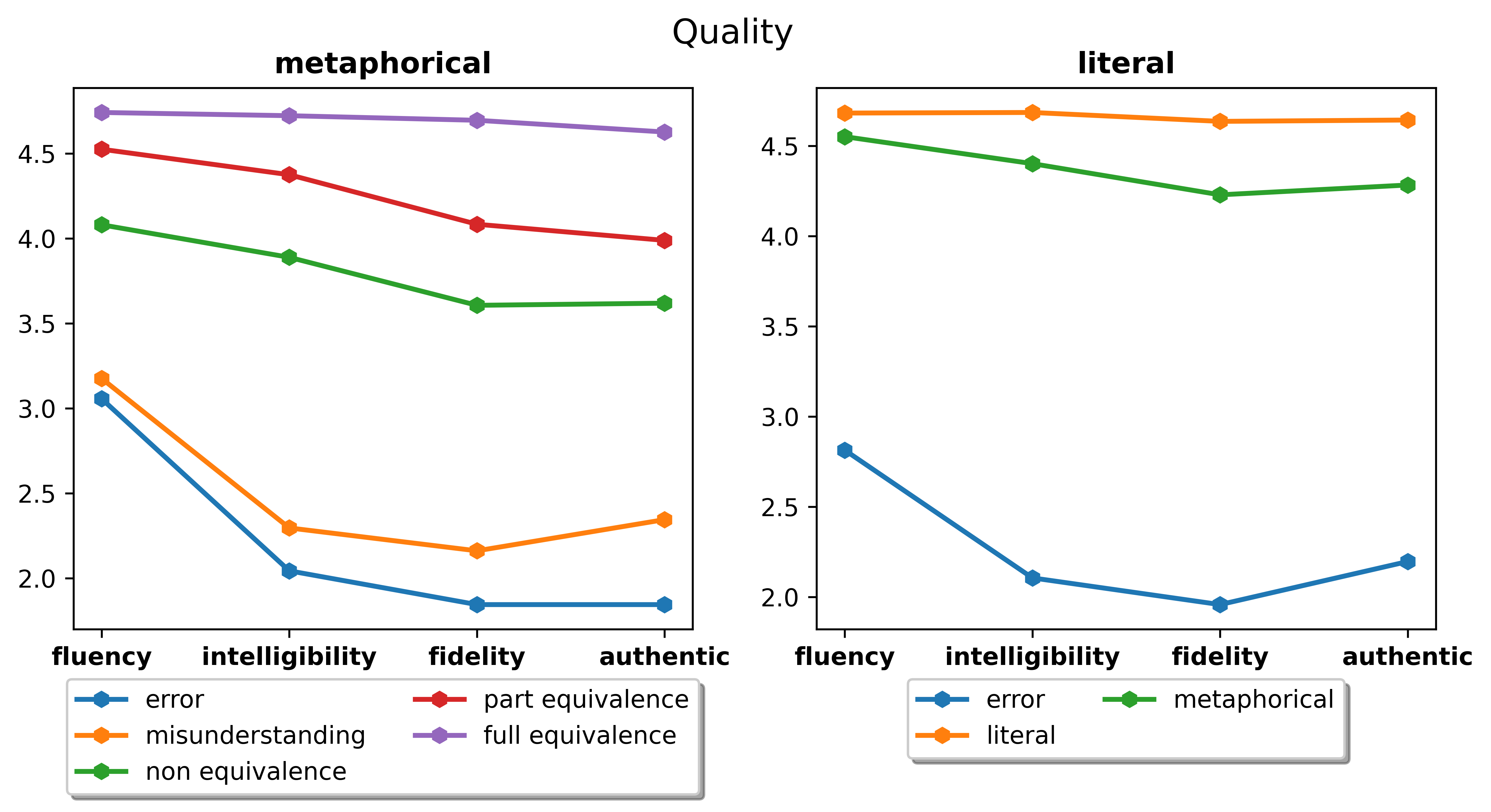}
    \caption{Average quality scores of manual evaluation of metaphorical and literal expression translation.}
    \label{fig:quality_equivalence}
\end{figure}
Besides maintaining emotional salience, fully equivalent metaphor translations and literal translations of literal expressions demonstrate higher translation quality. This is revealed in \autoref{fig:quality_equivalence}, which shows that fully equivalent translations of metaphorical expressions outperform others in the dimensions of Fluency, Intelligibility, Fidelity, and Authenticity, whilst literal translations of literal expressions outperform other versions in the four dimensions.
We also observe that part-equivalent and non-equivalent translations of metaphors cause more severe translation quality degradation than metaphorical translations of literal expressions. We hypothesise that literal translations of metaphorical expressions between languages spoken by different communities result in unnatural literal statements. which also supports the observation that the \textit{translation of metaphorical expressions is harder than that of literal expressions}.

\subsection{Impact of Different Language}
\label{sec:language_distance}

\begin{figure}[!h]
    \centering
    \includegraphics[width=\columnwidth]{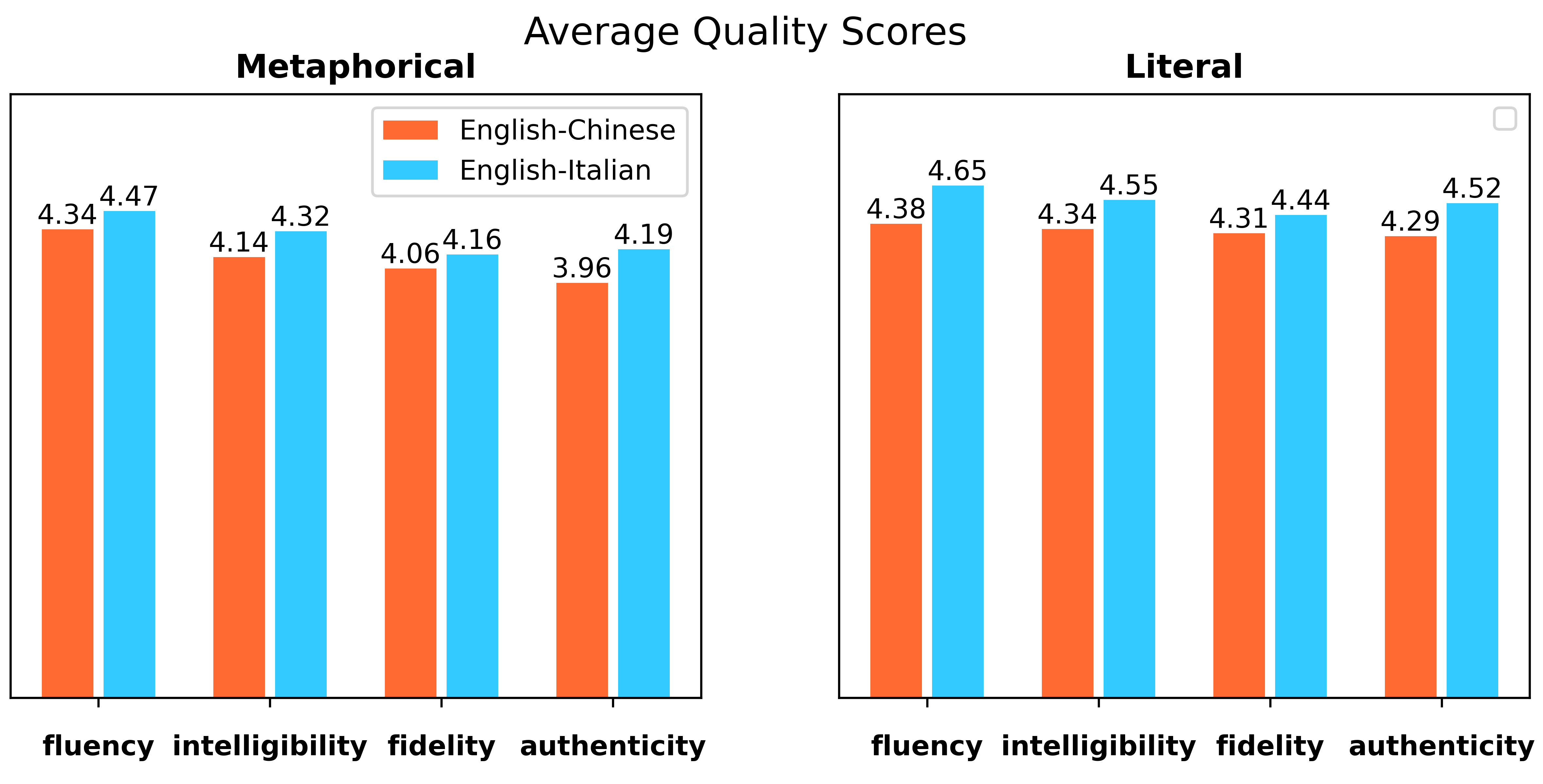}
    \caption{Average quality scores of manual evaluation for EN-ZH and EN-IT.}
    \label{fig:quality_comparison}
\end{figure}

\autoref{fig:quality_comparison} presents a comparison of the average evaluation scores of EN-ZH and EN-IT translations across all models. The results show that the average translation quality is lower for Chinese compared to Italian, despite both being translated from English. This can be attributed to several factors. Firstly, Chinese and English belong to different language families and possess distinct linguistic structures, with the grammatical disparities posing challenges for accurate translation. Secondly, cultural differences also play a significant role in translation quality. Translating metaphors accurately requires a deep understanding of cultural nuances and idiomatic expressions between the source and target languages. Failure to grasp these nuances can lead to mistranslation or loss of the intended meaning. Furthermore, the availability and quality of language resources and machine translation models differs for Chinese and Italian.

\section{Conclusion}

MMTE is the first work to systematically investigate how translations are affected by metaphor in a fine-grained and multi-lingual setting.
MMTE also introduces Equivalence as a new dimension of metaphor translation evaluation and verifies its relationship with emotional salience and translation quality.
Moreover, we conducted thorough experiments on the proposed evaluation dimensions and verified the increased difficulty of translating metaphorical expressions compared to literal expressions.
We further release MMTE, a high-quality metaphor translation corpus which can be adopted for automatic metric design for metaphor translation.
Future work intends to combine MMTE with additional well-designed automatic metrics aligning with specific human evaluation dimensions proposed in the paper.
\section{Limitations}
We summarise several limitations of MMTE which can be explored in future works.
Firstly, MMTE only conducts experiments on commercial state-of-the-art translation systems and the most well-known open-source translation packages, rather than models from research works. 
Secondly, due to resource scarcity for Italian language models and reliable Italian and Chinese metaphor detection models, we only provide thoughts on designing automatic metrics of metaphor translation evaluation based on our corpus, which we will release, rather than presenting plug-and-play automatic metrics.
Thirdly, we do not explore language typology in depth in Appendix \ref{Sec: Language Genetic} as it is an interesting side observation of MMTE. Additionally, it is only our working hypothesis that parallel corpus size is more critical for metaphor translation quality than linguistic typology, rather than a verified conclusion.

\section{Ethics Statement}
In conducting this research, we adhered to the highest ethical standards to ensure the integrity and responsibility of our work. The data used in our study were sourced from publicly available datasets, and no private or sensitive information was included. All human annotators involved in the study were fully informed about the research objectives and provided their consent prior to participation.

We ensured that the annotations and evaluations were conducted with fairness and respect for linguistic and cultural diversity. Additionally, the use of large language models (LLMs) was guided by ethical considerations, ensuring that the models were applied responsibly and their outputs were critically evaluated.

Our research aims to contribute positively to the field of computational linguistics by improving the quality and reliability of machine translation, particularly in the nuanced area of figurative language. We are committed to transparency, and our methodologies and findings are shared openly for peer review and further research.

\subsubsection*{Acknowledgments}
Tyler Loakman is supported by the Centre for Doctoral Training in Speech and Language Technologies (SLT) and their Applications funded by UK Research and Innovation [grant number EP/S023062/1].

\bibliography{anthology,custom}

\appendix
\section{Framework Details}
\label{Sec:framework}
\subsection{Translators and Languages}
\label{Sec:preprocess}
The \textbf{Google Cloud Translation API} (Translation V3 API) is a prominent commercial multilingual translation tool employing neural MT (NMT) techniques, known for its wide-ranging capabilities and comprehensive language support.

The \textbf{Youdao Cloud Translation API} is a popular commercial multilingual NMT tool within the Chinese community, proficient in handling Chinese language translation tasks.

The \textbf{Helsinki-NLP/opus-mt} models are pre-trained on the open parallel corpus (OPUS), a continuously expanding collection of translated texts sourced from the web. These models are widely used by researchers and practitioners due to their effectiveness and versatility.

The \textbf{GPT-4o}, developed by OpenAI, is an advanced language model designed to perform a wide range of natural language processing tasks, including serving as a highly capable translator model that can handle multiple languages with high accuracy and fluency.

\textbf{Chinese}, a Sino-Tibetan language, is renowned for its rich idiomatic expressions and extensive use of metaphors. \textbf{Italian}, a Romance language descended from Latin and belonging to the Indo-European language family like English, provides a distinct comparison. The distinction between these target languages enables a more accurate assessment of the models' performance in preserving metaphorical meaning.

\subsection{Annotation Setup}
\label{Sec:annotation_setup}
Our annotation platform is built on a private server using an open-source annotation tool - Doccano~\cite{doccano}.
We hired 18 annotators who are native speakers of the target languages, all of whom are linguistics majors with professional working competency in English. All annotation workers are paid based on the median wage of similar tasks on Amazon Turk, which is 10 dollars/hour.
Specifically, the annotators are divided into six groups, each with three annotators. 
The groups are further equally divided between annotating English-Chinese (\textbf{EN-ZH}) instances and English-Italian (\textbf{EN-IT}) instances.
Each group is tasked with labelling target words and post-editing the entire MOH dataset and its translations with all 647 pairs of data, resulting in each paired instance being annotated three times. 
In the final step, all annotation results are cross-checked by professional translators. 
A group meeting was held to discuss instances of disagreement, and final decisions were recorded on an online discussion website for future reference.
\section{Guideline}
\label{Sec:guideline}
\begin{CJK*}{UTF8}{gbsn}

In each annotation sample, an \textbf{English} sentence is be given as source text, followed by four translations in \textbf{Chinese}. Please evaluate each translation based on the criteria listed below. You will also be asked to supply your own translation as the gold reference.
\subsection{Sentence Quality}
Please compare the source sentence and its translation without reference to the correct translation, and evaluate the translation from following aspects:

\begin{itemize}
    \item \textbf{\textit{Fluency}}: To what extent the translation is well-formed and grammatical, ensuring that it sounds like it was originally written in the target language.
    \item \textbf{\textit{Intelligibility}}: To what extent the translation is easily understood and conveys metaphorical meaning sufficiently, such that readers can gain the intended interpretation.
    \item \textbf{\textit{Fidelity}}: The extent to which the translation is faithful to the source sentence, such that there is minimal distortion, twisting, or altering of meaning.
    \item \textbf{Overall}: An overall assessment to indicate the quality of the entire sentence seen as a whole.
\end{itemize}
Please judge these four aspects of quality on a 5-point Likert scale: 5) Very Good; 4) Good; 3) Acceptable; 2) Poor; 1) Very Poor.

\subsection{Equivalence}
Please compare the source sentence and its translation, and determine to what extent the target word and its translation are Equivalent in figuration. Here are the definitions of the three types of Equivalence and two types of mistakes:
\begin{itemize}
    \item \textbf{Full-Equivalence}: When comparing the source sentence and translation sentence, both the literal meanings and the contextual meanings of the target word are the same.\\
    \textbf{EN}: He @injected@ new life into the performance.\\	 \textbf{ZH}: 他给表演注入了新的生命
    \item \textbf{Part-Equivalence}: When comparing the source sentence and translation sentence, only the contextual meanings of the target word are similar. The literal meaning of the target word between the source sentence and translation are different, but they are both metaphorical.\\
    \textbf{EN}: @Wallow@ in your success!\\	 
    \textbf{ZH}: @沉浸@在你的成功中吧！
    \item \textbf{Non-Equivalence}: When comparing the source and translation, only the contextual meanings of the target word are similar. However, the translation is a non-metaphorical expression, making the literal meaning of the target word between the source and translation different.\\
    \textbf{EN}: Sales were @climbing@ after prices were lowered.\\	 
    \textbf{ZH}: 价格下跌后销售额@上升@。
    \item \textbf{Misunderstanding} When the literal meanings are similar between the target word in the source text and the target word in the translation, but the translation conveys no contextual meaning like the target in the source language.\\
    \textbf{EN}: I @attacked@ the problem as soon as I got out of bed.\\
    \textbf{ZH}: 我一下床就@攻击@了问题
    \item \textbf{Error}: When the target word is mistranslated, meaning that not only the contextual meanings are different between the source and translation, but their literal meanings also differ.\\
    \textbf{EN}: @Stamp@ fruit extract the juice.\\ 
    \textbf{ZH}: @果果@提取果汁。
    
\end{itemize}

\begin{figure}[!h]
    \centering
    \begin{subfigure}{1.0\columnwidth}
        \centering
        \includegraphics[width=\columnwidth]{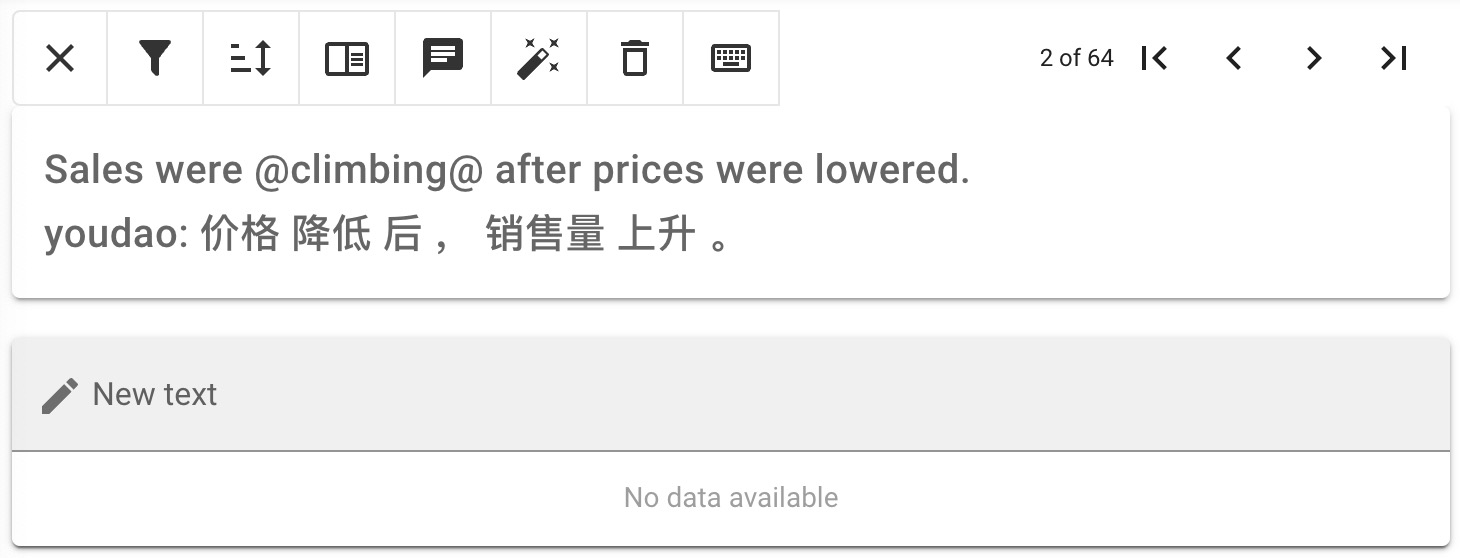}
        \caption{unlabeled sample}
    \end{subfigure}
    \begin{subfigure}{1.0\columnwidth}
        \centering
        \includegraphics[width=\columnwidth]{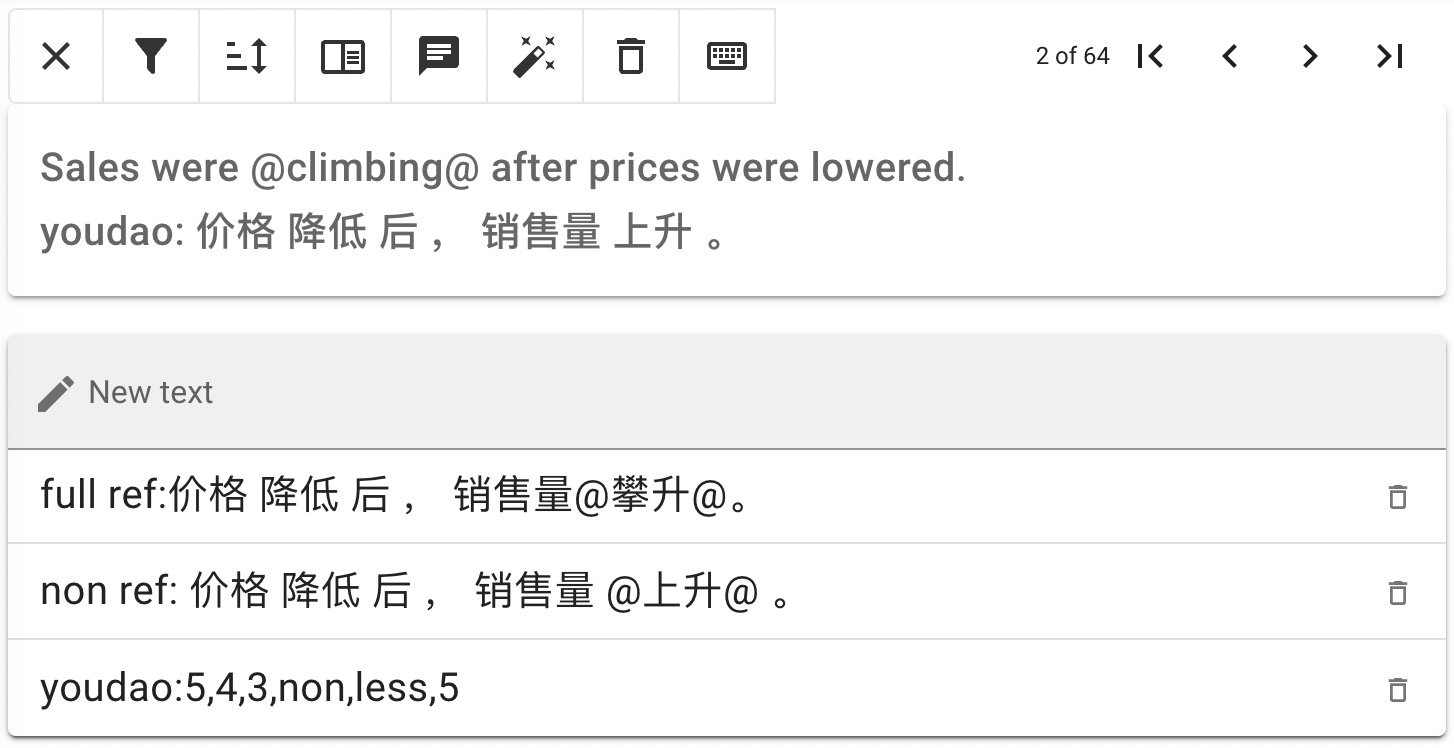}
        \caption{labeled sample}
    \end{subfigure}
    \caption{An example of the annotation process.}
    \label{fig:my_label}
\end{figure}
\end{CJK*}

\subsection{Emotion}
Please compare the source sentence and its translation, and determine to what extent the target word and its translation convey equal amounts of emotion. There are four labels to judge emotion:
\begin{itemize}
    \item \textbf{Zero}: If the target word in source context conveys no emotion, please fill \textbf{Zero}.\\
    \textbf{EN}: I can not @digest@ milk products.\\
    \textbf{ZH}: 我不能消化牛奶产品。
    \item \textbf{More}: The target word in the translation conveys \textbf{\textit{more}} emotion than the target word in the source sentence.\\
    \textbf{EN}: The seamstress @ruffled@ the curtain fabric.\\
    \textbf{ZH}: 裁缝女把窗帘布弄得一团糟.
    \item \textbf{Same}: The target words in the two sentences convey a \textbf{\textit{similar}} degree of emotion.\\
    \textbf{EN}: I @salute@ your courage!\\
    \textbf{ZH}: 我向你的勇气致敬!
    \item \textbf{Less}: The target word in the translation conveys \textbf{\textit{less}} emotion than the target word in the source sentence.\\
    \textbf{EN}: The spaceship blazed out into space. \\
    \textbf{ZH}: 太空船飞向太空
\end{itemize}

\subsection{Authenticity Target}
Please compare the target in the source sentence and its translation, and evaluate whether the target translation is authentic. In other words, to what extent is the translation idiomatic (i.e. is expressed in a way that a native speaker would express it)? 
Please judge the target on a 5-point scale: 5) Very Good; 4) Good; 3) Acceptable; 2) Poor; 1) Very Poor.

\subsection{Post-Editing}
By referring to the source sentence and its translations, in addition to the above Equivalence scale, please give two fluent and high-quality translations: 1) using figurative language (full-equivalence, part-equivalence) and 2) without using figurative language (non-equivalence). You should focus on the given target word, and make sure it is translated into an appropriate expression.

\section{Influence of Linguistic Typology on Translation Difficulty}
\label{Sec: Language Genetic}
Linguistic typological features are known to be able to assist translation and rank candidates for multilingual transfer \cite{oncevay-etal-2020-bridging}. 
The experimental results of Opus in \autoref{tab: comprehensive_evaluation} support a similar conclusion, that translation between a language pair with a closer typological relationship (EN-IT) is easier than a more distant pair (EN-ZH).
However, this conclusion does not hold for the experimental results of Google and Youdao in \autoref{tab: comprehensive_evaluation}.
Youdao, a popular commercial multilingual translation tool in the Chinese community, achieves better translation performance in the EN-ZH direction than EN-IT.
We hypothesize that the size of the corpus is much more important for translation quality compared to linguistic typology. Due to the above observations, the potentially larger EN-ZH parallel corpus that Youdao and Google have compared to EN-IT, and the relatively balanced sizes of EN-ZH and EN-IT parallel corpora that Opus holds, may aid in explaining the observed difference.

\section{Metaphor Explanation with LLMs}
\label{Sec:A_Explanation}

\begin{table*}[htb]
\resizebox{\textwidth}{!}{
    \centering
    \begin{tabular}{lp{16cm}}
    \toprule
    \textbf{Prompts}        & Please compare the source sentence and its translation, and determine to what extent the @target word@ and its translation are Equivalent in figuration. Here are the definitions of the three types of Equivalence and two types of mistake:
    Full-Equivalence: When comparing the source sentence and translation sentence, both the literal meanings and the contextual meanings of the target word are the same.
    Example: source: He @injected@ new life into the performance. translation: 他给表演@注入@了新的生命
    Part-Equivalence: When comparing between the source sentence and translation sentence, only the contextual meanings of the target word are similar. The literal meaning of the target word between the source sentence and translation are different, but they are both metaphorical.
    Example: source: @Wallow@ in your success! translation: @沉浸@在你的成功中吧！
    Non-Equivalence: When comparing the source and translation, only the contextual meanings of the target word are similar. However, the translation is a non-metaphorical expression, making the literal meaning of the target word between the source and translation different.
    Example: source: Sales were @climbing@ after prices were lowered. translation: 价格下跌后销售额@上升@。
    Misunderstanding: When the literal meanings are similar between the target word in the source text and the target word in the translation, but the translation conveys no contextual meaning like the target in the source language.
    Example: source: I @attacked@ the problem as soon as I got out of bed. translation: 我一下床就@攻击@了问题
    Error: When the target word is mistranslated, meaning not only that the contextual meanings are different between the source and translation, but their literal meanings also differ.
    Example: source: The @stamp@ fruit extract the juice. translation: 果果提取果汁。 \\ \hline
    \textbf{Query} & Source: \textbf{English source sentence} \newline Translation: \textbf{Target language translation} \newline Equivalence and Explanation:  \\
    \bottomrule
    \end{tabular}
}
\caption{Prompt and Query Format for LLMs in Evaluating Metaphor Translation Equivalence and Generating Explanations.}
\label{table:llm_prompt}
\end{table*}

\begin{table*}[htb]
\resizebox{\textwidth}{!}{
\centering
\begin{tabular}{lp{8cm}p{8cm}}
\toprule
   & \textbf{EN}: The news were  \underline{leaked} to the paper.             & \textbf{EN}: The colors don't \underline{harmonize.} \\
              & \textbf{ZH}: 消息 \underline{泄露} 给 了 报纸 。 & \textbf{IT}: I colori non si \underline{armonizzano}.\\
\hline
\textbf{GPT-3.5}        & Both the source sentence and the translation use a metaphorical expression to convey the same meaning. The target words "\underline{leaked}" and "\underline{泄露}" both literally and contextually mean that information was disclosed or released unintentionally or secretly.&  The source sentence and the translation both use the word "\underline{harmonize}" (and its Italian equivalent "\underline{armonizzano}") in the same context, with both the literal and contextual meanings being the same.\\ \hline
\textbf{GPT-4}          & The literal and contextual meanings of "\underline{leaked}" in the source sentence and "\underline{泄露}" in the translation are the same. Both terms refer to the unauthorized release or disclosure of information.&  The source sentence and the translation have both the same literal and contextual meanings of the target word "\underline{harmonize}". In both languages, it refers to the act of being in harmony or agreement. \\ \hline
\textbf{Gemini Pro} & **Literal Meaning:** Both "\underline{leaked}" and "\underline{泄露}" share the literal meaning of confidential information being unintentionally or secretly revealed. **Contextual Meaning:** Both words fit the context of the sentence, implying an act of disclosing information without authorization. The information flow (news to the paper) is consistent. Therefore, both the literal and contextual meanings are preserved in the translation, making it a case of Full-Equivalence.  & The word "\underline{harmonize}" and its Italian translation "\underline{armonizzano}" share both literal and contextual meanings. * **Literal Meaning:** Both words refer to the concept of elements coming together in a pleasing and balanced way.* **Contextual Meaning:** In the context of colours, both words imply that the colors used together do not create a visually appealing or cohesive effect. Therefore, the translation accurately conveys the meaning of the source word, making it a case of full equivalence. \\
\bottomrule
\end{tabular}
}
\caption{Examples of explanations given by the LLMs for metaphor translations, selecting one full-equivalence translation each in Chinese and Italian.}
\label{table:explanations}
\end{table*}

We used LLMs to annotate metaphor equivalence and attempted to guide the models to provide explanations for their evaluations. The specific formats of the prompts and queries are shown in the Tab.~\ref{table:llm_prompt}. By providing specific examples and explanations for each type of equivalence and including them in the prompts, we aimed to give LLMs references for comparison.

Given the powerful capabilities of large language models (LLMs), we employed LLMs to annotate and explain different metaphor translations. LLMs, particularly GPT-4, demonstrate an understanding that approaches human annotators in terms of both semantic and rhetorical comprehension.

As shown in Tab.~\ref{table:explanations}, LLMs showcased a robust ability to understand and interpret metaphors, providing comprehensive explanations that covered both semantic nuances and rhetorical aspects. This performance indicated a high level of competency in handling cross-linguistic tasks.

By analyzing the explanations provided by LLMs, we were able to validate their effectiveness in metaphor translation tasks. This analysis demonstrated that LLMs could not only understand and interpret metaphors accurately but also articulate the reasoning behind their evaluations. This capability is crucial for ensuring that the subtleties of metaphorical language are preserved in translation. LLMs offered reliable and insightful evaluations that are essential for high-quality translation assessment. Their ability to generate detailed and contextually accurate explanations for their decisions highlights their potential as a robust tool in the translation process.

\end{CJK*}
\end{document}